\documentclass[journal]{IEEEtran}
\ifCLASSINFOpdf
  \usepackage[pdftex]{graphicx}
  \graphicspath{{./figs/}}
\else
 \usepackage[dvips]{graphicx}
 \graphicspath{{./figs/}}
\fi

\usepackage{amsmath,graphicx,amssymb,amsthm,amsfonts,mathrsfs,cite}
\usepackage{algpseudocode,algorithm}
\usepackage{soul}
\usepackage{color}
\usepackage[normalem]{ulem}

\long\def\symbolfootnote[#1]#2{\begingroup%
	\def\thefootnote{\fnsymbol{footnote}}\footnote[#1]{#2}\endgroup} 

\title{On the Adversarial Robustness of LASSO Based Feature Selection}
\author{\IEEEauthorblockN{Fuwei Li, Lifeng Lai, and Shuguang Cui}}

\begin{document}
%
\maketitle
\symbolfootnote[0]{
	Fuwei Li and Lifeng Lai are with the Department of Electrical and Computer Engineering, University of California, Davis, CA, 95616 (e-mail: fli@ucdavis.edu; lflai@ucdavis.edu). 
	Shuguang Cui is currently with the Shenzhen Research Institute of Big Data and Future Network of Intelligence Institute (FNii), the Chinese University of Hong Kong, Shenzhen, China, 518172, and was with the Department of Electrical and Computer Engineering, University of California, Davis, CA, USA, 95616 (e-mail: shuguangcui@cuhk.edu.cn).
    The work was partially supported by the National Science Foundation with Grants CNS-1824553, CCF-1717943, and CCF-1908258.
	This paper was presented in part at IEEE International Workshop on
Machine Learning for Signal Processing, Espoo, Finland, Sep. 2020\cite{li2020advfeature}.
}

\begin{abstract}
In this paper, we investigate the adversarial robustness of feature selection based on the $\ell_1$ regularized linear regression model, namely LASSO. In the considered model, there is a malicious adversary who can observe the whole dataset, and then will carefully modify the response values or the feature matrix in order to manipulate the selected features. We formulate the modification strategy of the adversary as a bi-level optimization problem. Due to the difficulty of the non-differentiability  of the $\ell_1$ norm at the zero point, we reformulate the $\ell_1$
norm regularizer as linear inequality constraints. We employ the interior-point method to solve this reformulated LASSO problem and obtain the gradient information. Then we use the projected gradient descent method to design the modification strategy. In addition, We demonstrate that this method can be extended to other $\ell_1$ based feature selection methods, such as group LASSO and sparse group LASSO. Numerical examples with synthetic and real data illustrate that our method is efficient and effective. 
\end{abstract}
\begin{IEEEkeywords}
Linear regression, feature selection, LASSO, adversarial machine learning, bi-level optimization.
\end{IEEEkeywords}
\section{Introduction}
\label{sec:intro}
Feature selection is one of the most important pre-processing steps in the vast majority of machine learning and signal processing problems~\cite{dash1997feature,Mandanas2020feature,Furlane2006feature}. By performing feature selection, we can discard irrelevant and redundant features while keeping the most informative features. With the features of a smaller dimension, we can overcome the curse of dimensionality, better interpret our model, and speed up training and testing processes. Among a variety of feature selection methods, LASSO is one of the most widely used~\cite{tibshirani1996regression,tan2015lasso}. By putting the $\ell_1$ sparsity induced regularizer on a linear regression model, LASSO can perform feature selection and regression simultaneously. Owing to its simplicity and efficiency, LASSO is widely applied to bio-science~\cite{butcher2015probe}, financial analysis~\cite{zhang2019forecasting}, image processing~\cite{yang2017group}, etc. Furthermore, by exploring the additional structures of the regression coefficients, various extensions such as group LASSO~\cite{yuan2006model,lv2011lasso} and sparse group LASSO~\cite{simon2013sparse,zhang2015sparsegrouplasso} are proposed in the literature. 

Machine learning algorithms are used in a wide variety of applications: virtual personal assistance, video surveillance, recommendation system, etc. Among them, some are security and safety critical, such as medical image analysis, autonomous drive, and high-frequency trading~\cite{goodfellow2018making,li2020adversarial, finlayson2019adversarial,goldblum2020adversarial}. Despite its ubiquitous usage, recent study reveals that many of them are very vulnerable to adversarial attacks~\cite{goodfellow6572explaining,balda2019adversarial,xiao2015feature}. Since feature selection serves as the first stage of many of the machine learning algorithms, it is necessary and urgent to investigate its adversarial robustness. Though some existing works examined the robustness of feature selection against dense noise and outliers~\cite{jeong2018effect,loh2011high}, its behaviour under the adversary attacks is unknown. By analyzing the attack strategy of the adversary, the goal of our paper is to provide a better understanding of the sensitivity of feature selection methods against this kind of attacks.

In the considered feature selection model, we assume that there is an adversary who has the full knowledge of the model and can observe the whole dataset. After inspecting the dataset, it will carefully modify the response values or the feature matrix so as to manipulate the regression coefficients. By modifying the regression coefficients, it will maneuver the selected features. It can select the features which will not be selected originally by enlarging the magnitude of the corresponding regression coefficients. Also, it can make us wrongly discard important features by suppressing the magnitude of the corresponding regression coefficients. Moreover, it will try to make other regression coefficients unchanged so as to minimize the possibility of being detected by the feature selection system. In this paper, we intend to find the best modification strategy of the adversary with the energy constraints on the modification. By doing so, we can better understand how the response values and feature matrix influence the selected features and the robustness of the feature selection algorithm. 

We formulate this problem as a bi-level optimization problem.  The upper-level objective is to minimize the difference between the targeted regression coefficients and that learned from the modified dataset. The lower-level problem is just a LASSO based feature selection problem with the modified dataset.
To solve this bi-level optimization problem, we first solve the lower-level problem. Since the LASSO problem is a convex optimization problem, it is equivalent to its Karush–Kuhn–Tucker (KKT) conditions. By applying the implicit function theorem on the KKT conditions, we may learn the relationship between the dataset and the regression coefficients if the KKT conditions are continuous differentiable around its optimum. However, the $\ell_1$ norm is not continuous at the point zero. This prevents us from directly employing the implicit function theorem on the KKT conditions. To resolve the issue, we reformulate the LASSO problem as a linear inequality constrained quadratic programming problem and use the interior-point method to solve it. By utilizing the KKT conditions from the reformulated problem, we are able to find the gradients of our objective with respect to the response values and feature matrix. With the gradients information, we employ the projected gradient descent to solve this bi-level optimization problem. Similar methods can be applied to design the attack strategy based on the group LASSO and the sparse group LASSO.

Compared with~\cite{mei2015using} that requires the KKT conditions being continuous differentiable, our methods can be applied to any $\ell_1$ based feature selection methods. As we will show later, we successfully apply the proposed method to investigate the adversarial robustness of LASSO, group LASSO, and sparse group LASSO. Numerical examples with synthetic data and real data demonstrate that our method is efficient and effective. Our results show that feature selection based on the $\ell_1$ regularizer is very vulnerable to this kind of attacks. 

In this paper, we extensively expand upon the conference version of this work \cite{li2020advfeature}. Firstly, we extend the work of adversarial attack strategy against ordinary LASSO in \cite{li2020advfeature} to the strategy against group LASSO, which is widely used to analyze signals such as speech and image that have group-wise sparsity structure. Secondly, along with the attack against the group LASSO, in this paper we also demonstrate how to use our method to design the strategy to attack the feature selection algorithm based on sparse group LASSO, which is also widely used in medical analysis and text classification~\cite{simon2013sparse,vincent2014sparse}. We furthermore provide more comprehensive experiments to demonstrate our attack strategy in real applications, for example, direction of arrival estimation and weather prediction. 

The remainder of the paper is organized as follows. In Section~\ref{sec:prob-form}, we describe the precise problem formulation based on the ordinary LASSO feature selection method. 
In Section~\ref{sec:prob-ana}, we introduce our method to solve this problem.
In Section~\ref{sec:group-lasso}, we extend our method to attack the group LASSO and the sparse group LASSO based feature selection methods. 
In Section~\ref{sec:num-exams}, we provide comprehensive numerical experiments
with both synthetic data and real data to illustrate the results obtained in this paper. Finally, we offer concluding remarks in Section~\ref{sec:conclude}.

\section{Problem formulation}\label{sec:prob-form}
In this section we provide the problem formulation of adversarial attack against the ordinary LASSO based feature selection. 

Given the data set $\{(y_0^i,\, \mathbf{x}_0^i)\}_{i=1}^n$, where $n$ is the number of data samples, $y_0^i$ is the response value of data sample $i$, $\mathbf{x}_0^i \in \mathbb{R}^m$ denotes the feature vector of data sample $i$, and each element of $\mathbf{x}_0^i$ is called a feature of the data sample. Through the data samples, we attempt to learn a sparse representation of the response values from the features. The LASSO algorithm learns a sparse regression coefficient, $\boldsymbol{\beta}_0$, by solving
\begin{align}\label{beta0:lasso}
    \boldsymbol{\beta}_0 = \underset{\boldsymbol{\beta}}{\text{argmin}} \, \|\mathbf{y}_0-\mathbf{X}_0\boldsymbol{\beta} \|_2^2 + \lambda \|\boldsymbol{\beta}\|_1,
\end{align}
where the response vector $\mathbf{y}_0=[y_0^1,\,y_0^2,\dots,y_0^n]^\top$, the feature matrix $\mathbf{X}_0=[\mathbf{x}_0^1,\,\mathbf{x}_0^2,\dots,\mathbf{x}_0^n]^\top$, $\|\cdot\|_1$ denotes the $\ell_1$ norm, and $\lambda$ is the trade-off parameter to determine the relative goodness of fitting and sparsity of $\boldsymbol{\beta}_0$ \cite{tibshirani1996regression}. 
The locations of the non-zero elements of the sparse regression coefficients indicate the corresponding selected features.  

In this paper, we assume that there is an adversary who is trying to manipulate the learned regression coefficients, and thus maneuver the selected features by carefully modifying the response values or the feature matrix. We denote the modified response value vector as $\mathbf{y}$ and denote the modified feature matrix as $\mathbf{X}$. Further, we assume that the adversary's modification is constrained by the $\ell_p$ norm ($p \ge 1$). This means we have $\|\mathbf{y}-\mathbf{y}_0\|_p\le \eta_y$, and $\|\mathbf{X}- \mathbf{X}_0\|_p \le \eta_x$, where $\eta_y$ is the energy budget for the modification of the response values, and $\eta_{x}$ is the energy budget for the modification of the feature matrix. For a vector, $\|\cdot\|_p$ denotes the $\ell_p$ norm of the vector; for a matrix, $\|\cdot\|_p$ denotes the $\ell_p$ norm of the vectorization of the matrix. As a result, the manipulated regression coefficients, $\hat{\boldsymbol{\beta}}$, are learned from the modified data set $(\mathbf{y}, \mathbf{X})$ by solving the following LASSO problem 
\begin{align}\label{lasso}
    \hat{\boldsymbol{\beta}} =  \underset{\boldsymbol{\beta}}{\text{argmin}}\, \|\mathbf{y}-\mathbf{X}\boldsymbol{\beta} \|_2^2 + \lambda \|\boldsymbol{\beta}\|_1.
\end{align}

The goal of the adversary is to suppress or promote some of the regression coefficients while keeping the change of the remaining coefficients to be minimum. If it wants to suppress the $i$th regression coefficient, we minimize $s_i\cdot\hat{\beta}_i^2$, where ${s_i>0}$ is the predefined weight parameter. If it aims to promote the $i$th regression coefficient, we minimize $e_i\cdot\hat{\beta}_i^2$, where $e_i<0$ is the weight parameter.
To make the changes to the $i$th regression coefficient as small as possible, we minimize $\mu_i\cdot(\hat{\beta}_i - \beta_0^i)^2$, where $\mu_i >0$ is a user defined parameter to measure how much effort we put on keeping the $i$th regression coefficients intact. Moreover, we denote the set of indices of coefficients which are suppressed, promoted, and not changed as $S$, $E$, and $U$, respectively. In summary, the objective of the adversary is:
\begin{align}
    \min \, \frac{1}{2}(\hat{\boldsymbol{\beta}}- \boldsymbol{\nu})^\top \mathbf{H} (\hat{\boldsymbol{\beta}}- \boldsymbol{\nu}),
    \label{prob:obj}
\end{align}
where $\nu_i=\beta_0^i$ if $i\in U$, otherwise $\nu_i=0$, and $\mathbf{H}=\text{diag}(\mathbf{h})$ as $h_i=\mu_i$ for $i\in U$, $h_i=s_i$ for $i\in S$ and $h_i=e_i$ for $i\in E$. 

Considering the energy constraints of the adversary, to obtain the optimal attack strategy, we need to solve a bi-level optimization problem:
\begin{align}\label{opt:prob-orig}
  \underset{\mathbf{y} \in \mathcal{C}_y, \mathbf{X}\in \mathcal{C}_x}{\min}&\quad  \frac{1}{2}
  (\hat{\boldsymbol{\beta}}- \boldsymbol{\nu})^\top \mathbf{H} (\hat{\boldsymbol{\beta}}- \boldsymbol{\nu}) \\
  \text{s.t.}&\quad  \hat{\boldsymbol{\beta}} = \underset{\boldsymbol{\beta}}{\text{argmin}}\, 
  \|\mathbf{y}-\mathbf{X}\boldsymbol{\beta} \|_2^2 + \lambda \|\boldsymbol{\beta}\|_1,
  \label{opt:lower-level}
\end{align}
where $\mathcal{C}_y =\left\{\mathbf{y}\,|\, \|\mathbf{y}-\mathbf{y}_0\|_p\le \eta_y \right\}$ and \\
$\mathcal{C}_x = \left\{\mathbf{X}\, |\, \| \mathbf{X}-\mathbf{X}_0\|_p \le \eta_x \right\}$. 

\section{Algorithm}\label{sec:prob-ana}
In this section, we investigate problem~\eqref{opt:prob-orig} and present our projected gradient descent method to solve this problem. 

In problem~\eqref{opt:prob-orig}, the objective is a function of $\hat{{\boldsymbol{\beta}}}$. However, the relationship between $(\mathbf{y},\mathbf{X})$ and $\hat{\boldsymbol{\beta}}$ is determined by the lower-level optimization problem. This makes our objective a very complicated function of $(\mathbf{y},\mathbf{X})$. 
So, the bi-level optimization problem is difficult to solve in general. To solve this bi-level optimization problem, we need to first solve the lower-level optimization problem to determine the dependence between $(\mathbf{y}, \mathbf{X})$ and $\hat{\boldsymbol{\beta}}$. Then, we can use the gradient descent method to solve this bi-level optimization problem. 
Since the lower-level problem is convex \cite{tibshirani1996regression}, it can be represented by its KKT conditions. The corresponding KKT conditions with respect to the lower-level optimization problem is: 
\begin{align}\label{kkt:l1}
\mathbf{0} \in 2\mathbf{X}^\top(\mathbf{X}\boldsymbol{\beta}-\mathbf{y}) + \lambda \partial \|\boldsymbol{\beta}\|_1,
\end{align} 
where, $\partial\|\cdot\|_1$ is the subgradient of the $\ell_1$ norm. We denote the right hand of~\eqref{kkt:l1} as $q(\boldsymbol{\beta}, \mathbf{y}, \mathbf{X})$. 

If $q(\boldsymbol{\beta}, \mathbf{y}, \mathbf{X})$ is a continuous differentiable function and its Jacobian matrix with respect to $\boldsymbol{\beta}$ is invertible, the KKT conditions define a one-to-one mapping from $(\mathbf{y}, \mathbf{X})$ to $\boldsymbol{\beta}$, and by the implicit function theorem~\cite{dontchev2009implicit},  we can calculate the gradient of $\boldsymbol{\beta}$ with respect to $\mathbf{y}$ and $\mathbf{X}$.  
Unfortunately, in our case, $q(\boldsymbol{\beta}, \mathbf{y}, \mathbf{X})$ is not differentiable at the point with $\beta_i=0$. Moreover, \eqref{opt:lower-level} does not always determine a single valued mapping from $(\mathbf{y}, \mathbf{X})$ to $\boldsymbol{\beta}$. For example, when $\lambda \ge \|\mathbf{X}^\top \mathbf{y}\|_\infty$, we always have $\boldsymbol{\beta}=\mathbf{0}$. 

To circumvent these difficulties, we transform the lower-level optimization problem to the following equivalent linear inequality constrained quadratic programming \cite{kim2007interior}:
\begin{align}\label{opt:lasso-linear}
    \underset{\boldsymbol{\beta}, \mathbf{u}}{\text{argmin}}&\quad \|\mathbf{y} - \mathbf{X}\boldsymbol{\beta}\|_2^2 + \lambda \sum_{i=1}^m u_i \\
    \text{s.t.}&\quad -u_i \le \beta_i \le u_i,\, i=1,2,\dots,m,
\end{align}
where $\mathbf{u}=[u_1,u_2,\dots,u_m]^\top$. 
Following~\cite{kim2007interior}, we can apply the interior-point method to solve~\eqref{opt:lasso-linear}. In particular, we solve the penalized problem:
\begin{align}\label{opt:lasso-penalty}
    \underset{\boldsymbol{\beta}, \mathbf{u}}{\text{argmin}} \quad  \|\mathbf{y} - \mathbf{X}\boldsymbol{\beta}\|_2^2 + \lambda \sum_{i=1}^m u_i + \frac{1}{t}\Phi(\boldsymbol{\beta},\mathbf{u}),
\end{align}
where $\Phi(\boldsymbol{\beta},\mathbf{u})=-\sum_{i=1}^m \log(u_i^2-\beta_i^2)$ is the penalty function for the constraints of \eqref{opt:lasso-linear} and $t$ is the penalty parameter. Solution of problem~\eqref{opt:lasso-penalty} converges to~\eqref{lasso} if we follow the central path as $t$ varies from $0$ to $\infty$. 

Instead of using the KKT conditions of ~\eqref{kkt:l1}, we utilize the KKT conditions of~\eqref{opt:lasso-penalty}, which are 
\begin{align}\label{kkt:pen1}
    2\mathbf{X}^\top(\mathbf{X}\boldsymbol{\beta}-\mathbf{y})+
    \frac{1}{t}
    \begin{bmatrix}
    2\beta_1/(u_1^2-\beta_1^2),\\
    \vdots\\
    2\beta_m/(u_m^2-\beta_m^2)
    \end{bmatrix}
    &= \mathbf{0}, \\
    \lambda \mathbf{1} - 
    \frac{1}{t}
    \begin{bmatrix}
    2u_1/(u_1^2 - \beta_1^2)\\
    \vdots\\
    2u_m/(u_m^2 - \beta_m^2 )
    \end{bmatrix}
    &=\mathbf{0}. 
    \label{kkt:pen2}
\end{align}
Let us denote the KKT conditions as $\mathbf{g}( \mathbf{y},\mathbf{X}, \boldsymbol{\beta}, \mathbf{u})=\mathbf{0}$. 
 According to the implicit function theorem,
the derivative of $\boldsymbol{\beta}$ with respect to $\mathbf{y}$ can be computed as the first $m$ rows of
\begin{align}
    -\mathbf{J}^{-1}\frac{\partial \mathbf{g}}{\partial \mathbf{y}},
    \label{exp:grad:beta-y}
\end{align}
where $\mathbf{J}=
[\frac{\partial \mathbf{g}}{\partial \boldsymbol{\beta}}, \frac{\partial \mathbf{g}}{\partial \mathbf{u}}]$ is the Jacobian matrix of $\mathbf{g}(\mathbf{y}, \mathbf{X}, \boldsymbol{\beta}, \mathbf{u})$ with respect to $\boldsymbol{\beta}$ and $\mathbf{u}$,
\begin{align}
    \frac{\partial \mathbf{g}}{\partial \mathbf{y}}
    &= 
    \begin{bmatrix}
      -2\mathbf{X}^\top \\
      \mathbf{0}
    \end{bmatrix},\\   
    \frac{\partial \mathbf{g}}{\partial \boldsymbol{\beta}} 
    &= 
    \begin{bmatrix}
    2\mathbf{X}^\top \mathbf{X} + \mathbf{D}_1\\
    \mathbf{D}_2
    \end{bmatrix}, \\
    \frac{\partial \mathbf{g}}{\partial \mathbf{u}}
    &=
    \begin{bmatrix}
    \mathbf{D}_2 \\
    \mathbf{D}_1
    \end{bmatrix},
\end{align}
with
\begin{alignat*}{2}
    \mathbf{D}_1 &=\frac{1}{t} \text{diag}
    \big(&&
    2(u_1^2+\beta_1^2)/(u_1^2-\beta_1^2)^2,\dots,\\
    & &&2(u_m^2+\beta_m^2)/(u_m^2-\beta_m^2)^2
    \big),\\
    \mathbf{D}_2&=\frac{1}{t} \text{diag}\big(
    && -4u_1\beta_1/(u_1^2-\beta_1^2)^2, \dots,\\ 
    & && -4u_m\beta_m/(u_m^2-\beta_m^2)^2\big). 
\end{alignat*}
Also, according to~\eqref{kkt:pen1}, ~\eqref{kkt:pen2}, and the implicit function theorem, 
the derivative of $\boldsymbol{\beta}$ with respect to $\mathbf{X}$ can be calculated as the first $m$ rows of 
\begin{align}
    -\mathbf{J}^{-1}\frac{\partial \mathbf{g}}{\partial \mathbf{X}},
    \label{exp:grad:beta-x}
\end{align}
where
$\frac{\partial \mathbf{g}}{\partial \mathbf{X}} \in \mathbb{R}^{2m\times(mn)}$ with
\begin{align}
    \frac{\partial g_i}{\partial X_{kl}}
    = 
    \begin{cases}
    2\delta_{li}(\mathbf{X}\boldsymbol{\beta}-\mathbf{y})_k + 2X_{ki}\beta_l, &\mbox{if }i\le m \\
    0, & \mbox{if } i >m
    \end{cases}
\end{align}
with $\delta_{li}$ being the Kronecker delta function 
\begin{align*}
    \delta_{li}
    =
    \begin{cases}
    1, &\mbox{if } i=l, \\
    0, &\mbox{if } i\neq l,
    \end{cases}
\end{align*}
and $(\mathbf{X}\boldsymbol{\beta}-\mathbf{y})_k$ being the $k$th element of the vector $(\mathbf{X}\boldsymbol{\beta}-\mathbf{y})$.  

To calculate the gradient of $\boldsymbol{\beta}$ with respect to $\mathbf{y}$ and $\mathbf{X}$, we first need to find the inverse of the Jacobian matrix. The Jacobian matrix is a $2\times2$ block matrix,
\begin{align*}
    \mathbf{J} =
    \begin{bmatrix}
    2\mathbf{X}^\top \mathbf{X}+\mathbf{D}_1 & \mathbf{D}_2 \\
    \mathbf{D}_2 &\mathbf{D}_1
    \end{bmatrix}. 
\end{align*}
This block structure makes the inverse of $\mathbf{J}$ admit a simple form \cite{lu2002inverses}:
\begin{align}
    \mathbf{J}^{-1} =
    \begin{bmatrix}
    (2\mathbf{X}^\top\mathbf{X}+2\mathbf{D})^{-1} & * \\
    * & * 
    \end{bmatrix}, 
    \label{eq:inv-jacob}
\end{align}
where $\mathbf{D}=1/t\cdot\text{diag}\big(1/(u_1^2+\beta_1^2), \dots, 1/(u_m^2+\beta_m^2) \big)$. Since the gradients of $\boldsymbol{\beta}$ with respect to $\mathbf{y}$ and $\mathbf{X}$ only depend on the first $m$ rows of~\eqref{exp:grad:beta-y} and~\eqref{exp:grad:beta-x} respectively and the elements from $m+1$ to $2m$ are zero both for $\partial \mathbf{g} / \partial \mathbf{y}$ and $\partial \mathbf{g}/\partial\mathbf{X}$, we omit unused elements in~\eqref{eq:inv-jacob} in later computation. 
With this explicit expression of the Jacobian matrix, we have
\begin{align}\label{dbdy}
    \frac{\partial \boldsymbol{\beta}}{\partial \mathbf{y}} = \big(\mathbf{X}^\top\mathbf{X}+\mathbf{D}\big)^{-1}\mathbf{X}^\top, 
\end{align} 
and 
\begin{align}
    \frac{\partial \boldsymbol{\beta}}{\partial X_{kl}} =
    \left[\frac{\partial \beta_1}{\partial X_{kl}}, \frac{\partial \beta_2}{\partial X_{kl}}, \dots, 
    \frac{\partial \beta_m }{\partial X_{kl}} \right]^\top ,
\end{align}
with 
$$\frac{\partial \beta_i }{\partial X_{kl}} =\sum_{j} -(\mathbf{X}^\top\mathbf{X} +  \mathbf{D})_{ij}^{-1}\frac{\partial g_j}{\partial X_{kl}}.$$

\begin{algorithm}[t!]
\caption{The Projected Gradient Descent Algorithm}\label{alg:pgd}
\begin{algorithmic}[1]
\State \textbf{Input}: data set $\{(y_0^i, \mathbf{x}_0^i)\}_{i=1}^n$, trade off parameter $\lambda$ in~\eqref{beta0:lasso}, energy budget $\eta_y$, $\eta_x$, $\ell_p$ norm, and step-size parameter $\alpha_t$.

\State solve $\boldsymbol{\beta}_0$ via~\eqref{beta0:lasso}, set up feature sets $S$, $E$, $U$ and their corresponding parameters $\mathbf s$, $\mathbf e$, $\boldsymbol{\mu}$; use those parameters to define the objective function $f(\mathbf{y},\mathbf{X})$ in~\eqref{opt:prob-orig}.

\State \textbf{Initialize} set  the number of iterations $t=0$ and $\mathbf{y}_t = \mathbf{y}_0$, $\mathbf{X}_t = \mathbf{X}_0$. 

\State \textbf{Do}

\State solve $\hat{\boldsymbol{\beta}}$ according to \eqref{opt:lasso-penalty}, 

\State compute the gradients: $\nabla_{\mathbf{y}} f(\mathbf{y}_t, \mathbf{X}_t)$ according to~\eqref{grad:y} and 
$\nabla_{\mathbf{X}}f(\mathbf{y}_t,\mathbf{X}_t)$ according to~\eqref{grad:X},
\State update: \\
$\mathbf{y}_{t+1} = (1-\alpha_t)\mathbf{y}_t + \alpha_t\cdot\text{Proj}_{\mathcal{C}_y}\big(\mathbf{y}_t -\nabla_{\mathbf{y}} f(\mathbf{y}_t, \mathbf{X}_t)\big),$
\State update:\\
$\mathbf{X}_{t+1} = (1-\alpha_t)\mathbf{X}_t + \alpha_t\cdot\text{Proj}_{\mathcal{C}_x}\big(\mathbf{X}_t -\nabla_{\mathbf{X}} f(\mathbf{y}_t, \mathbf{X}_t)\big),$
\State set $t = t+1$,
\State \textbf{While} convergence conditions are not met.
\State \textbf{Output}: $\mathbf{y}_t, \mathbf{X}_t$.
\end{algorithmic}
\end{algorithm}

Let us denote the objective of~\eqref{opt:prob-orig} as $f(\mathbf{y},\mathbf{X})$. Using the chain rule, we have the gradient of $f$ with respect to $\mathbf{y}$ and $\mathbf{X}$:
\begin{align}
    \nabla_{\mathbf{y}}f(\mathbf{y},\mathbf{X}) = 
    \left(
    \frac{\partial \boldsymbol{\beta}}{\partial \mathbf{y}}\right)^\top 
    \mathbf{H}(\boldsymbol{\beta}- \boldsymbol{\nu})\Big|_{\boldsymbol{\beta}=\hat{\boldsymbol{\beta}}}
    \label{grad:y}
\end{align}
and
\begin{align}
    \frac{\partial f(\mathbf{y}, \mathbf{X})}{\partial X_{kl}} = (\boldsymbol{\beta}-\boldsymbol{\nu})^\top\mathbf{H}\frac{\partial \boldsymbol{\beta}}{\partial X_{kl}}\Big|_{\boldsymbol{\beta} = \hat{\boldsymbol{\beta}}}.
    \label{grad:X}
\end{align}

Now, we know the gradients of our objective function~\eqref{opt:prob-orig}. With the help of this gradient information, we can use a variety of gradient based optimization methods. Since our problem is a constrained optimization problem, we resort to the projected gradient descent method. We have summarized it in Algorithm~\ref{alg:pgd}. The main concept of the projected gradient descent algorithm is that we first take a gradient step, project it onto the feasible set, and then take an $\alpha_t$ step toward the projected point. In this algorithm, $\text{Proj}_{\mathcal{C}_y}(\cdot)$ and $\text{Proj}_{\mathcal{C}_x}(\cdot)$ represent the projection operators that project a point onto the feasible set $\mathcal{C}_y$ and $\mathcal{C}_x$, respectively. $\mathcal{C}_y$ and $\mathcal{C}_x$ are $\ell_p$ balls with radius $\eta_y$ and $\eta_x$ respectively. 
In the following, we will discuss the expressions of the projection onto three commonly used $\ell_p$ norm balls, where $p=1,2, \infty$ with the radius of the norm ball being $\eta$ and its center being the origin. 
 
\noindent
\textbf{Case 1}: Project onto the $\ell_1$ norm ball. It involves solving the following convex problem
\begin{align*}
    \underset{\mathbf{z}}{\text{argmin}} &\quad \|\mathbf z- \mathbf x\|_2 \\
    \text{s.t.}&\quad \|\mathbf{z}\|_1\le \eta, 
\end{align*}
where $\mathbf{x}$ is the point to be projected. It can be efficiently solved via its dual with complexity $\mathcal{O}(m)$ \cite{condat2016fast}. 
\\
\noindent
\textbf{Case 2}: Project onto the $\ell_2$ norm ball. In this case, we have a very simple closed form solution
\begin{align}
    \text{Proj}(\mathbf{x})=\mathbf{x}./\max\{1, \|\mathbf{x}\|_2/\eta\},
\end{align}
where `$./$' denotes the element-wise division. 
\\
\noindent
\textbf{Case 3}: Project onto the $\ell_\infty$ norm ball. In this case, we also have a very simple closed-form solution:
\begin{align}
    \text{Proj}(\mathbf{x})= \mathbf{z},
\end{align}
where $\mathbf{z}=[z_1,\dots,z_m]^\top$ and
\begin{align*}
    z_i = 
    \begin{cases}
    -1, &\mbox{if } x_i \le -\eta, \\
    x_i, &\mbox{if } |x_i| <\eta, \\
    1,  &\mbox{if } x_i \ge \eta. 
    \end{cases}
\end{align*}
With these expressions of the projection, we can easily obtain the expressions of $\text{Proj}_{\mathcal{C}_y}(\cdot)$ and $\text{Proj}_{\mathcal{C}_x}(\cdot)$ by simply doing geometric translation.

\section{Adversarial Attacks against  Group LASSO and Sparse Group LASSO}\label{sec:group-lasso}
In this section, we will extend the method developed in Section~\ref{sec:prob-ana} to design optimal attack strategy towards two other popular LASSO based feature selection methods: group LASSO and sparse group LASSO. 

\subsection{Adversarial Attacks Against Group LASSO}
Many of the sparse signals such as speech signal \cite{chen2014group} and functional brain network \cite{zhao2018functional,ziniel2013dynamic}, possess additional special structures. To select the most useful features, it is better to exploit these additional structures \cite{simon2013sparse}. 
The group LASSO imposes a group-wise sparsity structure, i.e., only a few groups have nonzero entries. 
This group-wise sparsity guides us to select better features,
such as in splice site detection~\cite{roth2008group} and hyperspectral image classification~\cite{yang2017group}.
The group-wise sparsity structure can be promoted by solving the following group LASSO problem:
\begin{align}
    &\min_{\boldsymbol{\beta}} \quad \left\|\mathbf{y} - \sum_{l=1}^L\mathbf{X}_l \boldsymbol{\beta}_l\right\|_2^2 + 
    \lambda \sum_{l=1}^L\sqrt{p_l}\|\boldsymbol{\beta}_l\|_2.
    \label{group-lasso}
\end{align}
Here the feature matrix $\mathbf{X}$ is divided into $L$ groups, each of which $\mathbf{X}_l\in \mathbb{R}^{n\times p_l}$, $\sum_{l=1}^L p_l = m$, and $\boldsymbol{\beta}=[\boldsymbol{\beta}_1^\top,\boldsymbol{\beta}_2^\top,\dots,\boldsymbol{\beta}_L^\top]^\top$. The regularization term $\lambda\sum_{l=1}^L\sqrt{p_l}\|\boldsymbol{\beta}_l\|_2$ is used to promote the group-wise sparse structure, and $\lambda$ is the penalty parameter to control the sparsity level and goodness of fitting. 

Considering our attack target and the energy budget constraints for modifying the response values and the feature matrix, the design of optimal feature manipulation attacks for the group LASSO can be cast as a bi-level optimization:
\begin{align}\nonumber 
    \min_{\mathbf{y}\in \mathcal{C}_y, \mathbf{X}\in \mathcal{C}_x}
    &\quad \frac{1}{2}(\hat{\boldsymbol{\beta}}-\boldsymbol{\nu})^\top \mathbf{H}(\hat{\boldsymbol{\beta}}-\boldsymbol{\nu}) \\
    \text{s.t.}\quad & 
    \hat{\boldsymbol{\beta}} = \underset{\boldsymbol{\beta}}{\text{argmin}}\quad 
    \left\| 
    \mathbf{y} - \sum_{l=1}^L\mathbf{X}_l\boldsymbol{\beta}_l
    \right\|_2^2 
    +\lambda\sum_{l=1}^L\sqrt{p_l}\|\boldsymbol{\beta}_l\|_2,
\end{align}
where $\boldsymbol{\nu}$ and $\mathbf{H}$ are defined the same as in problem~\eqref{prob:obj}.

To solve this bi-level optimization problem, we also first consider the lower-level group LASSO problem. The group LASSO is a convex optimization problem, which is equivalent to the following quadratic programming with conic constraints:
\begin{align}
    \underset{\boldsymbol{\beta},\,\boldsymbol{\alpha}}{\text{argmin}} \quad 
    & \left\|\mathbf{y} - \sum_{l=1}^L \mathbf{X}_l\boldsymbol{\beta}_l\right\|_2^2 + \sum_{l=1}^L \lambda_l \alpha_l \\\nonumber
    \text{s.t.}\quad
    & \|\boldsymbol{\beta}_l\|_2 \le \alpha_l, \quad l=1,2,\dots, L, 
    \label{group-lasso-to-conic}
\end{align}
where $\lambda_l = \lambda\sqrt{p_l}$ and $\boldsymbol{\alpha} = [\alpha_1,\alpha_2,\dots,\alpha_L]^\top$. To solve this problem, we can utilize the similar interior-point method we have employed for the ordinary LASSO problem in Section~\ref{sec:prob-ana}. In particular, we solve a series of the minimization problems: $\min\, f_t$, as $t$ gradually grows, where
\begin{align*}
    f_t = 
    \left\|\mathbf{y} - \sum_{l=1}^L\mathbf{X}_l\boldsymbol{\beta}_l\right\|_2^2 + 
    \sum_{l=1}^L\lambda_l\alpha_l -
    1/t\sum_{l=1}^L \log(\alpha_l^2 - \|\boldsymbol{\beta}_l\|_2^2).
\end{align*}
Since this interior-point objective $f_t$ is a convex function, the minimization problem is equal to its KKT conditions:
\begin{align}\nonumber
    \nabla_{\boldsymbol{\beta}_l} f_t 
    &= \mathbf{X}_l^\top\left(\sum_{l=1}^L \mathbf{X}_l\boldsymbol{\beta}_l - \mathbf{y}\right) + \frac{1}{t}\frac{1}{\alpha_l^2 - \|\boldsymbol{\beta}_l\|_2^2}\boldsymbol{\beta}_l = \mathbf{0}, \\ \nonumber
    \frac{\partial f_t}{\partial \alpha_l} 
    &= \lambda_l -\frac{2}{t}\frac{\alpha_l}{\alpha_l^2 - \|\boldsymbol{\beta}_l\|_2^2} = \mathbf{0}, \\\nonumber
    &\quad \text{ for } l=1, 2, \dots, L. 
\end{align}
To derive the gradients of $\boldsymbol{\beta}$ with respect to $\mathbf{y}$ and $\mathbf{X}$, we can apply the implicit function theorem on these KKT conditions. First, we need to compute the Jacobian matrix of the function on the left of the KKT conditions. The gradients of $\nabla_{\boldsymbol{\beta}} f_t$ with respect to $\boldsymbol{\beta}$ and $\boldsymbol{\alpha}$ can be computed by 
\begin{align*}
    \nabla_{\boldsymbol{\beta}_j}\nabla_{\boldsymbol{\beta}_i} f_t = 
    \begin{cases}
    2\mathbf{X}_i^\top \mathbf{X}_j, &\mbox{ for } i\neq j, \\
    2\mathbf{X}_i^\top \mathbf{X}_j + \frac{1}{t}\frac{(\alpha_i^2 - \boldsymbol{\beta}_i^\top \boldsymbol{\beta}_i)\mathbf{I}+2\boldsymbol{\beta}_i\boldsymbol{\beta}_i^\top}{(\alpha_i^2 - \boldsymbol{\beta}_i^\top\boldsymbol{\beta}_i)^2}, 
    &\mbox{ for } i=j, 
    \end{cases}
\end{align*}

\begin{align*}
    \frac{\partial}{\partial \alpha_j}\nabla_{\boldsymbol{\beta}_i} f_t = 
    \begin{cases}
    \mathbf{0}, &\mbox{ for } i\neq j, \\
    \frac{-4}{t}\frac{\alpha_i\boldsymbol{\beta_i}}{(\alpha_i^2 - \|\boldsymbol{\beta}_i\|_2^2)^2}, &\mbox{ for } i=j. 
    \end{cases}
\end{align*}
The gradients of $\nabla_{\boldsymbol{\alpha}}f_t$ with respect to $\boldsymbol{\beta}$ and $\boldsymbol{\alpha}$ can be computed by 
\begin{align*}
    \nabla_{\boldsymbol{\beta}_j}\frac{\partial f_t}{\partial \alpha_i}&= 
    \begin{cases}
    \mathbf{0},
    &\mbox{ for } i\neq j, \\
    \frac{-4}{t}\frac{\alpha_i\boldsymbol\beta_i}{(\alpha_i^2 - \|\boldsymbol{\beta}_i\|_2^2)^2}
    &\mbox{ for } i=j, 
    \end{cases} \\
    \frac{\partial^2 f_t}{\partial \alpha_i \partial \alpha_j} &=
    \begin{cases}
    0,
    &\mbox{ for } i\neq j, \\
    \frac{2}{t}
    \frac{\alpha_i^2+\boldsymbol{\beta}_i^\top \boldsymbol{\beta}_i}{(\alpha_i^2 - \|\boldsymbol{\beta}_i\|_2^2)^2}, 
    &\mbox{ for } i=j. 
    \end{cases}
\end{align*}
Hence, the Jacobian matrix is 
\begin{align*}
    \mathbf{J} = 
    \begin{bmatrix}
    &\nabla_{\boldsymbol{\beta}}\nabla_{\boldsymbol{\beta}} f_t &  \nabla_{\boldsymbol{\beta}}\nabla_{\boldsymbol{\alpha}} f_t\\ 
    &\nabla_{\boldsymbol{\alpha}} \nabla_{\boldsymbol{\beta}}f_t & 
    \nabla_{\boldsymbol{\alpha}}\nabla_{\boldsymbol{\alpha}} f_t
    \end{bmatrix}.
\end{align*}
Let $\mathbf{g} =[\nabla_{\boldsymbol{\beta}}^\top f_t, \nabla_{\boldsymbol{\alpha}}^\top f_t]^\top $. Then we have 
\begin{align*}
    \nabla_{\boldsymbol{y}}\mathbf{g} = 
    \begin{bmatrix}
    -2\mathbf{X}^\top\\
    \mathbf{0}
    \end{bmatrix},
\end{align*}
and 
\begin{align*}
    \frac{\partial g_k}{\partial X_{ij}}=
    \begin{cases}
    2\left[\delta_{kj}(\mathbf{X}\boldsymbol{\beta} - \mathbf{y})_i + X_{ik}y_j\right], 
    &\mbox{ for } 1\le k \le m, \\
    0, 
    &\mbox{ otherwise}. 
    \end{cases}
\end{align*}
As a result, the gradient of $\boldsymbol{\beta}$ with respect to $\mathbf{y}$ and $\mathbf{X}$ is the first $m$ rows of $-\mathbf{J}^{-1}\nabla_{\mathbf{y}}\mathbf{g}$ and $-\mathbf{J}^{-1}\frac{\partial \mathbf{g}}{\partial \mathbf{X}}$, respectively. 
With this gradient information and using the chain rule, we can obtain the gradients of our objective with respect to the response values and feature matrix. Then, we can use the projected gradient descent method described in Algorithm~\ref{alg:pgd} to design our attack strategy. 

\subsection{Adversarial Attacks Against Sparse Group LASSO}
Sparse group LASSO combine the ordinary and the group LASSO and exploit the sparsity and group sparsity jointly.  
By combining these two properties, sparse group LASSO promotes the group-wise sparsity as well as the sparsity within each group. By taking advantages of these two perspective sparsities, sparse group LASSO help us select more accurate features and it has been used in climate prediction~\cite{chatterjee2012sparse}, heterogeneous feature representations~\cite{zhao2015heterogeneous}, change-points estimation~\cite{zhang2015sparsegrouplasso}, etc.
The sparse group LASSO problem tries to solve the following convex problem:
\begin{align}
    \min_{\boldsymbol{\beta}}\quad &\|\mathbf{y} - \sum_{l=1}^L\mathbf{X}_l \boldsymbol{\beta}_l \|_2^2 
    +
    \lambda_1\sum_{l=1}^L\sqrt{p_l}\|\boldsymbol{\beta}_l\|_2
    +
    \lambda_2\|\boldsymbol{\beta}\|_1.
\end{align}
Similar to problem~\eqref{group-lasso}, we assume the regression coefficients are divided into $L$ groups and each group $\boldsymbol{\beta}_l \in \mathbb{R}^{p_l}$. In the above objective, the first term is the ordinary least square to measure the goodness of fitting, the second term promotes the group-wise sparsity, and the third term encourages the sparsity within each group. 

Taking objective \eqref{prob:obj} into account, the design of optimal attack strategy against sparse group LASSO can be formulated as solving a bi-level optimization problem: 
\begin{align}
    \min_{\mathbf{y}\in \mathcal{C}_y, \mathbf{X}\in \mathcal{C}_x}
    &\quad \frac{1}{2}(\hat{\boldsymbol{\beta}}-\boldsymbol{\nu})^\top \mathbf{H}(\hat{\boldsymbol{\beta}}-\boldsymbol{\nu}) \\\nonumber
    \text{s.t.}
    &\quad 
   \hat{\boldsymbol{\beta}}=\underset{\boldsymbol{\beta}}{\text{argmin}}\quad \|\mathbf{y} - \sum_{l=1}^L\mathbf{X}_l \boldsymbol{\beta}_l \|_2^2 \\\nonumber
    &\qquad \qquad \qquad +
    \lambda_1\sum_{l=1}^L\sqrt{p_l}\|\boldsymbol{\beta}_l\|_2
    +
    \lambda_2\|\boldsymbol{\beta}\|_1.
\end{align}

To solve this bi-level optimization problem, as in the previous subsection, we can transform the lower-level problem into a quadratic programming with conic and linear inequality constraints by introducing the new variables $\alpha_l$ for $l=1,2,\dots, L$ and $u_i$ for $i=1,2,\dots, m$ as follows:
\begin{align}
    \underset{\boldsymbol{\beta},\boldsymbol{\alpha},\mathbf{u} }{\text{argmin}}
    \quad 
    &\|\mathbf{y} - \sum_{l=1}^L \mathbf{X}_l \boldsymbol{\beta}_l\|_2^2 + \sum_{l=1}^L \Tilde{\lambda}_l \alpha_l + \lambda_2 \sum_{i=1}^m u_i \\
    \text{s.t.}\quad
    & \|\boldsymbol{\beta}\|_2 \le \alpha_l, \quad l = 1, 2, \dots, L, \\
    & -u_i \le \beta_i \le u_i,\quad i=1,2,\dots, m, 
    \label{prob:sg-lasso}
\end{align}
where $\Tilde{\lambda}_l = \lambda_1\sqrt{p}_l$. We use the similar interior-point method to solve this optimization problem. Thus, we use penalty functions for the constraints and have the new objective with a certain penalty parameter $t$:
\begin{align*}
    h_t =& \|\mathbf{y} - \sum_{l=1}^L\mathbf{X}_l\boldsymbol{\beta}_l\|_2^2 + \sum_{l=1}^L \Tilde{\lambda}_l \alpha_l + \lambda_2 \sum_{i=1}^m u_i \\
    &-1/t\sum_{l=1}^L\log(\alpha_l^2 - \|\boldsymbol{\beta}_l\|_2^2)
    -1/t\sum_{i=1}^m \log(u_i^2 - \beta_i^2).
\end{align*}
The corresponding KKT conditions are 
\begin{align*}
    \begin{cases}
    \nabla_{\boldsymbol{\beta}_l}h_t =
    &2\mathbf{X}_l^\top(\mathbf{X}\boldsymbol{\beta} - \mathbf{y}) + 1/t\cdot \frac{2\boldsymbol{\beta_l}}{\alpha_l^2 - \|\boldsymbol{\beta}_l\|_2^2} \\
    &+\frac{2\boldsymbol{\beta}_l}{t}\cdot
    \text{diag}\Big( 1/\left((u_l^1)^2 -( \beta_l^1\right)^2,\\
    &\dots, 1/\left((u_l^{p_l})^2 - (\beta_l^{p_l})^2 \right)\Big)=\boldsymbol{0}, \\
    &\mbox{for } l=1, 2, \dots, L, \\
    \frac{\partial h_t}{\partial \alpha_l} 
    =& \Tilde{\lambda}_l - 1/t\cdot\frac{2\alpha_l}{\alpha_l^2 - \|\boldsymbol{\beta}_l\|_2^2}=0, \mbox{ for } l=1, 2, \dots, L, \\
    \frac{\partial h_t}{\partial u_i} =
    &\lambda_2 - 1/t\cdot \frac{2u_i}{u_i^2 -\beta_i^2}=0, \mbox{ for } i=1,2,\cdots, m, 
    \end{cases}
\end{align*}
where $\boldsymbol{\beta} =[\boldsymbol{\beta}_1^\top,\boldsymbol{\beta}_2^\top,\dots,\boldsymbol{\beta}_L^\top]^\top$, $\mathbf{u}=[\mathbf{u}_1^\top,\mathbf{u}_2^\top,\dots,\mathbf{u}_L^\top]^\top$, $\boldsymbol{\beta}_l=[\beta_l^1, \beta_l^2,\dots,\beta_l^{p_l}]^\top$, $\mathbf{u}_l = [u_l^1, u_l^2, \dots, u_l^{p_l}]^\top$ and $\text{diag}\left(x_1, x_2, \dots, x_n \right)$ is the diagonal matrix with diagonal entries $[x_1, x_2, \dots, x_n]$. To use the implicit function theorem to obtain the gradient information, we need to compute the Jacobian matrix of the function on the left of KKT conditions. The Jacobian matrix is 
\begin{align*}
    \mathbf{J} = 
    \begin{bmatrix}
    &\nabla_{\boldsymbol{\beta}}\nabla_{\boldsymbol{\beta}} h_t & \nabla_{\boldsymbol{\alpha}}\nabla_{\boldsymbol{\beta}} h_t & \nabla_{\mathbf{u}}\nabla_{\boldsymbol{\beta}} h_t \\
    & \nabla_{\boldsymbol{\beta}}\nabla_{\boldsymbol{\alpha}}h_t & \nabla_{\boldsymbol{\alpha}}\nabla_{\boldsymbol{\alpha}}h_t & \nabla_{\mathbf{u}}\nabla_{\boldsymbol{\alpha}}h_t \\
    &\nabla_{\boldsymbol{\beta}}\nabla_{\mathbf{u}}h_t & \nabla_{\boldsymbol{\alpha}}\nabla_{\mathbf{u}}h_t & \nabla_{\mathbf{u}}\nabla_{\mathbf{u}}h_t
    \end{bmatrix},
\end{align*}
where 
\begin{align*}
    &\nabla_{\boldsymbol{\beta}}\nabla_{\boldsymbol{\beta}}h_t =
    2\mathbf{X}^\top \mathbf{X}+\mathbf E_{1,1} + \mathbf D_{1,1},
\end{align*}
in which
\begin{align*}
\mathbf E_{1,1} = \nonumber
&\frac{1}{t}\text{diag}\bigg( \frac{(\alpha_1^2 - \boldsymbol{\beta}_1^\top \boldsymbol{\beta}_1)\mathbf{I}+2\boldsymbol{\beta}_1\boldsymbol{\beta}_1^\top}{(\alpha_1^2 - \boldsymbol{\beta}_1^\top\boldsymbol{\beta}_1)^2}, 
\dots, \\ 
&\frac{(\alpha_L^2 - \boldsymbol{\beta}_L^\top \boldsymbol{\beta}_L)\mathbf{I}+2\boldsymbol{\beta}_L\boldsymbol{\beta}_L^\top}{(\alpha_L^2 - \boldsymbol{\beta}_L^\top\boldsymbol{\beta}_L)^2} 
\bigg),\\ \nonumber
\mathbf D_{1,1} =& 2/t\cdot\text{diag}\big((u_1^2 + \beta_1^2)/(u_1^2 - \beta_1^2)^2, \dots,\\ 
&(u_m^2+\beta_m^2)/(u_m^2 - \beta_m^2)^2 \big),
\end{align*}

\begin{align*}
    \frac{\partial}{\partial \alpha_j}\nabla_{\boldsymbol{\beta}_i} f_t = 
    \begin{cases}
    \mathbf{0}, &\mbox{ for } i\neq j, \\
    \frac{-4}{t}\frac{\alpha_i\boldsymbol{\beta_i}}{(\alpha_i^2 - \|\boldsymbol{\beta}_i\|_2^2)^2}, &\mbox{ for } i=j,
    \end{cases}
\end{align*}

\begin{align*} \nonumber
    \nabla_{\mathbf{u}}\nabla_{\boldsymbol{\beta}} h_t = 
    &\text{diag}\bigg(
    -4/t\cdot\frac{\beta_1 u_1}{(u_1^2-\beta_1^2)^2},\dots,\\
    &-4/t\cdot\frac{\beta_m u_m}{(u_m^2-\beta_m^2)^2}
    \bigg), 
\end{align*}

\begin{align*}
    \frac{\partial^2 f_t}{\partial \alpha_i \partial \alpha_j} &=
    \begin{cases}
    0,
    &\mbox{ for } i\neq j, \\
    \frac{2}{t}
    \frac{\alpha_i^2+\boldsymbol{\beta}_i^\top \boldsymbol{\beta}_i}{(\alpha_i^2 - \|\boldsymbol{\beta}_i\|_2^2)^2}, 
    &\mbox{ for } i=j,
    \end{cases}
\end{align*}
\begin{align*}
    \nabla_{\mathbf{u}}\nabla_{\boldsymbol{\alpha}}h_t = \mathbf{0}, 
\end{align*}
and 
\begin{align*} \nonumber
    \nabla_{\mathbf{u}}\nabla_{\mathbf{u}}h_t = &\text{diag}
    \Big(
    2(u_1^2+\beta_1^2)/(u_1^2-\beta_1^2)^2,\dots,\\ 
    & 2(u_m^2+\beta_m^2)/(u_m^2-\beta_m^2)^2
    \Big).
\end{align*}
Let $\mathbf{q} \triangleq[\nabla_{\boldsymbol{\beta}}h_t^\top, \nabla_{\boldsymbol{\alpha}} h_t^\top, \nabla_{\mathbf{u}} h_t ^\top]^\top$, then we have 
\begin{align*}
    \nabla_{\mathbf{y}}\mathbf{q} =
    \begin{bmatrix}
    -2\mathbf{X}^\top\\
    \mathbf{0}\\
    \mathbf{0}
    \end{bmatrix}
\end{align*}
and 
\begin{align*}
    \frac{\partial q_k}{\partial X_{ij}}=
    \begin{cases}
    2\left[\delta_{kj}(\mathbf{X}\boldsymbol{\beta} - \mathbf{y})_i + X_{ik}y_j\right], 
    &\mbox{ for } 1\le k \le m, \\
    0, 
    &\mbox{ otherwise}. 
    \end{cases}
\end{align*}
Then we have the gradient of $\boldsymbol{\beta}$ with respect to $\mathbf{y}$ being the first $m$ rows of 
$$-\mathbf{J}^{-1}\nabla_{\mathbf{y}}\mathbf{q}$$
and the partial derivative of $\beta_k$ with respect to $X_{i,j}$ is
$$\frac{\partial \beta_k}{\partial X_{i,j}} = \sum_{l=1}^m -(\mathbf{J}^{-1})_{k,l}\frac{\partial q_l}{\partial X_{i,j}}.$$

Having the gradients of $\boldsymbol{\beta}$ 
with respect to $\mathbf{y}$ and $\mathbf{X}$, combining the gradients of our objective with respect to $\boldsymbol{\beta}$ and using the chain rule, we can get the full gradients of our objective with respect to $\mathbf{y}$ and $\mathbf{X}$. With these gradients information, we can then employ the projected gradient descent described in Algorithm~\ref{alg:pgd} to find our modification strategy. 

\section{Numerical Examples}\label{sec:num-exams}
In this section, we carry out several experiments to demonstrate the results obtained in this paper. 

\subsection{Attack Against Ordinary LASSO}
In the first numerical example, we test our algorithm on a synthetic data set. Firstly, we generate a $30\times50$ feature matrix $\mathbf{X}_0$. Each entry of the feature matrix is i.i.d. generated from a standard normal distribution. Then, we generate the response values, $\mathbf{y}_0$, through the model $\mathbf{y}_0 = \mathbf{X}_0\mathbf{v}+\mathbf{n}$, where $\mathbf{v}$ is the sparse vector in which only ten randomly selected positions are non-zero and each of the non-zero entry is i.i.d. drawn from the standard normal distribution; $\mathbf{n}$ is the noise vector where each entry is i.i.d. generated according to a normal distribution with zero mean and $0.1$ variance. Then, we set the LASSO trade-off parameter $\lambda = 2$ and use~\eqref{opt:lasso-linear} to estimate the regression coefficients $\boldsymbol{\beta}_0$. We randomly select one regression coefficient as the desired coefficient to be boosted and another one as the coefficient to be suppressed. In addition, we set the suppressed parameter $s_i = 1$ for $i\in S$, set boosted parameter $e_i = -1$ for $i\in E$, and set the unchanged parameter $\mu_i =5$ for $i \in U$. We set the step-size parameter $\alpha_t = 2/\sqrt{t}$ in Algorithm~\ref{alg:pgd}. 

\begin{figure}[t]
    \centering
    \includegraphics[width=.7\linewidth]{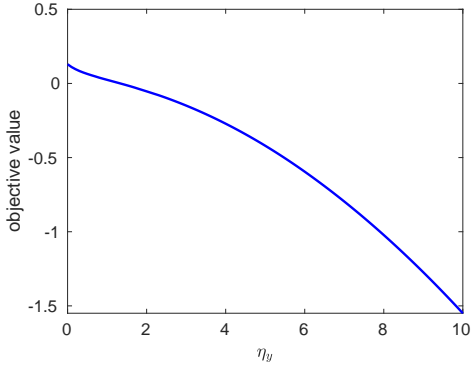}
    \caption{The objective value changes with the energy budget.}
    \label{fig:exp1:fval-eta}
\end{figure}
\begin{figure}
    \centering
    \includegraphics[width=.7\linewidth]{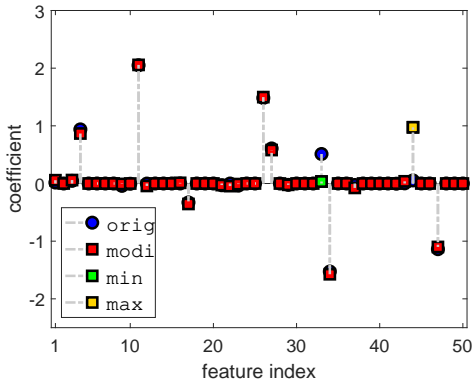}
    \caption{The original regression coefficients and the regression coefficients after our attacks. Here, `orig' denotes the original regression coefficients, `modi' represents the regression coefficients after our attack, `min' is the regression coefficient we want to suppress, and `max' denotes the regression coefficient we want to promote.}
    \label{fig:exp1:one-instance}
\end{figure}

In the first experiment, we set $\eta_x =0$, which means that we do not modify the feature matrix, and impose $\ell_2$ norm constraint on the modification of the response values. Then, we vary the energy budget, $\eta_y$, to see how energy budget influence our objective value. Fig.~\ref{fig:exp1:fval-eta} illustrates that the objective value decreases as the energy budget increases, which is expected as a larger energy budget provides a larger feasible region, and thus lower objective value. Fig.~\ref{fig:exp1:one-instance} demonstrates the recovered regression coefficients when $\eta_y=5$ along with the original regression coefficients. As the figure demonstrates, we have successfully suppressed and promoted the corresponding coefficients while keeping other regression coefficients almost unchanged. 

\begin{figure}[t!]
    \centering
    \includegraphics[width=.8\linewidth]{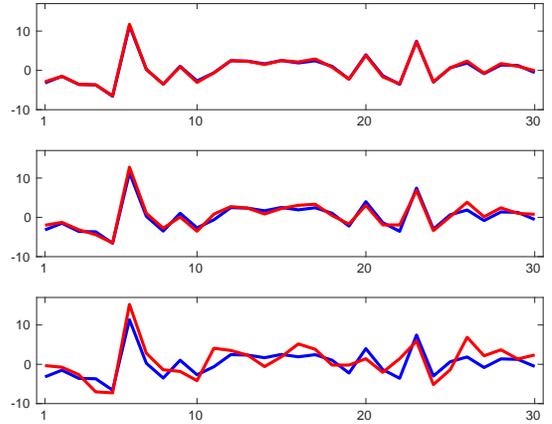}
    \caption{The blue line demonstrates the original response values and the red line is the modified response values with different attack constraints. From top to bottom are the modified response values with $\ell_1$, $\ell_2$, and $\ell_\infty$ norm constraints, respectively.}
    \label{fig:exp2:ylps}
\end{figure}
In the second experiment, we also attack the response values. We fix the energy budget $\eta_y =5$ and test different $\ell_p$ norm constraints on the modification of the response values as $p=1,2,\infty$. Fig.~\ref{fig:exp2:ylps} shows the original and modified response values under different $\ell_p$ norm constraints. The $x$-axis denotes the index of each response value and the $y$-axis denotes the value of the response vector. From the figure, we can see that the $\ell_1$ norm constraint provides the smallest modification on the response values and the $\ell_\infty$ norm constraint provides the most significant modification, which results in objective value $0.0095$ with the $\ell_1$ norm constraint, objective value $-0.4199$ with the $\ell_2$ norm constraint, and objective value $-2.8813$ with the $\ell_\infty$ norm constraint. That is because with the same radius, $\ell_1$ norm ball is contained in the $\ell_2$ norm ball and $\ell_2$ norm ball belongs to the $\ell_\infty$ norm ball. 

In the third experiment, we compare the modification on the response values and on the feature matrix with the $\ell_1$ constraints. First, we only attack the response values with $\eta_y=5$, which results in objective value $0.0095$. Second, we only attack the feature matrix with the same energy budget $\eta_x =5$, which results in objective value $-0.0969$. Finally, we attack both the response values and the feature matrix with $\eta_y =5$ and $\eta_x =5$, which results in objective value $-0.2291$. These results indicate that both the modifications of the response values and feature matrix are effective.

\begin{figure}[t]
    \centering
    \includegraphics[width=.7\linewidth]{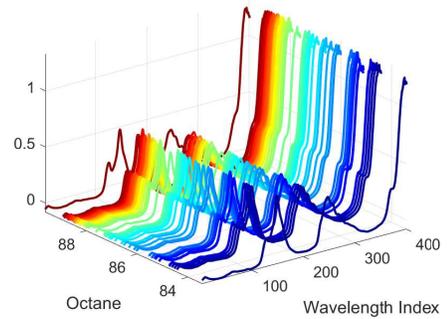}
    \caption{Overview of the octane data set.}
    \label{fig:octane-overview}
\end{figure}

\begin{figure}[t]
    \centering
    \includegraphics[width=.7\linewidth]{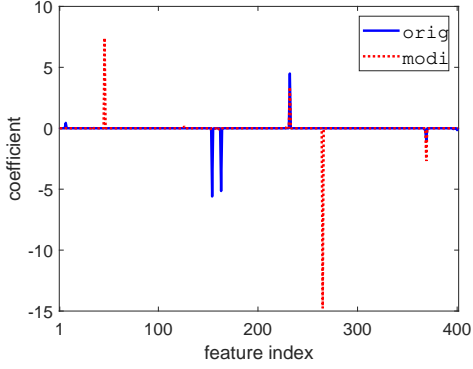}
    \caption{The regression coefficients before and after our attack.}
    \label{fig:octan-coeffs}
\end{figure}

We now test our attack strategy using real datasets. In this task, we use the spectral intensity of the gasoline to predict its octane rating \cite{kalivas1997two}. It consists of $60$ samples of gasoline at $401$ wavelength and their octane ratings. Fig.~\ref{fig:octane-overview} provides an overview of the data samples. In this figure, the octane axis indicates the octane rating of each sample and the z-axis denotes the spectral intensities at different wavelengths. From the figure we can see that there are very high correlations among different wavelengths. Hence, if we use the ordinary linear regression method, it will have large errors. Thus, we use the LASSO method to complete the regression task. We randomly choose $80\%$ of the data samples as our training data and the rest as our test data. We do cross-validation on the training data to decide the trade-off parameter in LASSO, and it gives $\lambda = 0.5$. Using this parameter, we compute the regression coefficients. Using this regression coefficients on the test data set, we have r-squared value $0.979$. The blue line in Fig.~\ref{fig:octan-coeffs} shows the original regression coefficient. From this figure, we can see that there are several important features.

In the next step, we modify the response values and the feature matrix with the energy budget $\eta_y = 5$ and $\eta_x=5$ to suppress the $154$th and $163$th regression coefficients, keep the $232$th and $369$th regression coefficients unchanged, and promote the rest of the regression coefficients. In our algorithm, we set $s_i=1$ for $i\in S$, $e_i=-1$ for $i\in E$, $\mu_i=50$ for $i\in U$, and step-size parameter $\alpha_t = 1/t$. The red-dashed line in Fig.~\ref{fig:octan-coeffs} shows the regression coefficients after our attacks. From the figure, we can see that we successfully promote two regression coefficients which were zero-valued before attack. We also suppress the $154$th and $163$th regression coefficients and make the $232$th and $369$th regression coefficients change very little. Using this regression coefficients on the test data set, we got the r-squared value $0.694$. Hence, by changing the response values and the feature matrix, we can easily make the system choose the wrong features. 

\subsection{Attack Against Group LASSO}

\begin{figure}[t!]
    \centering
    \includegraphics[width=0.7\linewidth]{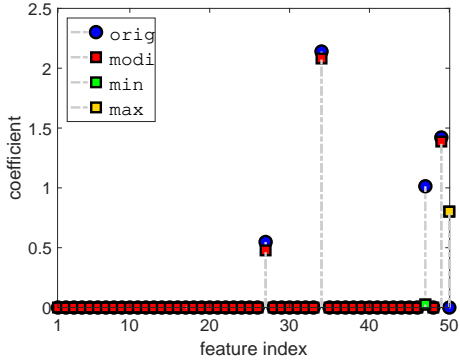}
    \caption{The magnitude of the coefficients before and after attacks. Here, `orig' denotes the original regression coefficients, `modi' represents the regression coefficients after attack,  `min' and `max' indicate the coefficients we want to supperss and boost after attack, respectively.}
    \label{fig:doa-coffs}
\end{figure}

\begin{figure}[t!]
\centering 
\begin{minipage}[b]{0.7\linewidth}
  \centering
  \centerline{\includegraphics[width=\linewidth]{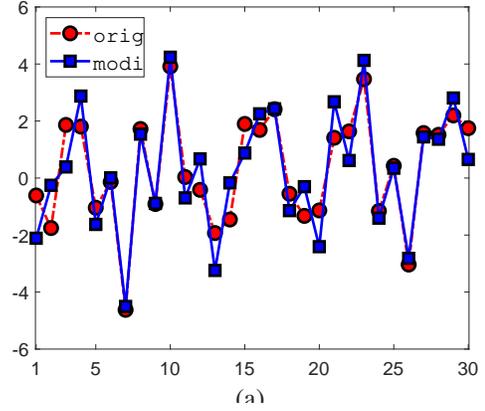}
  }

  \centerline{(a)}\medskip
\end{minipage}
\hfill
\begin{minipage}[b]{0.7\linewidth}
  \centering
  \centerline{\includegraphics[width=\linewidth]{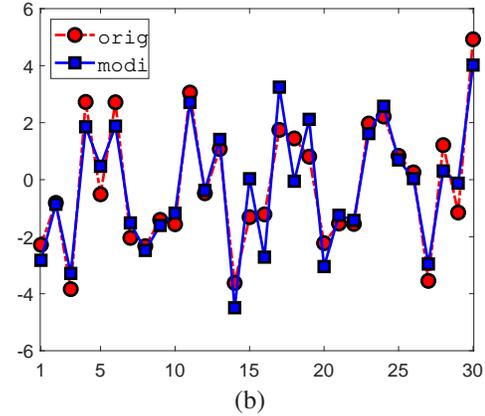}}
  \centerline{(b)}\medskip
\end{minipage}
\caption{(a) represents the real part of the observed signal and (b) the imaginary part of the observed signal before and after attacks. }
\label{fig:doa-y}
\end{figure}

In this subsection, we will employ our attack strategy on group LASSO. We will use the direction of arrival (DOA) problem as an example. 
In the DOA problem, we try to find the directions of the sources from the received signals of an array of sensors~\cite{stoica1990maximum,shan1985spatial}. 
Consider a setup where the sensors are linearly located and equally spaced with half of the wavelength. Hence, the measurement of the $n$th sensor are 
$\sum_{k=1}^K e^{j2\pi n f_k} x_k$, where $K$ is the number of sources and $f_k\in(-\pi/2,\pi/2]$ is the arrival angle of the $k$th source. Furthermore, we assume that the number of input sources is limited. If we divide the arrival of angle equally into $N$ grids and assume the sources are located on the grids, the DOA can be modeled as a linear signal acquisition system:
$$\mathbf{y} = \mathbf{A} \mathbf{x} + \mathbf{e},$$
where $\mathbf{y}\in \mathbb{C}^N$ is the measurements of the sensors, $\mathbf{A}\in \mathbb{C}^{N\times M}$,  $A_{n,m}= e^{j2\pi n\frac{m-1}{M}}$, $\mathbf{x}\in\mathbb{C}^{M}$ is the sparse source vector where only the locations which have targets are non-zero , and $\mathbf{e}\in\mathbb{C}^N$ is the noise vector.
We can first recover the sparse signal $\mathbf{x}$, then the arrival angles can be derived from the locations of the non-zero components of $\mathbf{x}$. Further, we can solve the following LASSO problem to recover $\mathbf{x}$:
\begin{align}
    \underset{\mathbf{x}}{\text{argmin}}:
    &\quad \|\mathbf{y} - \mathbf{A}\mathbf{x}\|_2^2 
    +\lambda \|\mathbf{x}\|_1,
    \label{prob:complex-lasso}
\end{align}
where the $\ell_1$ norm of $\mathbf{x}$ is defined as 
\begin{align}
    \|\mathbf{x}\|_1 = \sum_{i=1}^N \sqrt{(x^{R}_i)^2 + (x_i^I)^2},
\end{align}
and $x_i^R$ and $x_i^I$ are the real and imaginary parts of $x_i$, respectively. Problem~\eqref{prob:complex-lasso} is actually a group LASSO problem if we separate its real and imaginary parts and we reformulate it as:
\begin{align}
    \underset{\mathbf{x}^R, \mathbf{x}^I}{\text{argmin}}
     \quad \|\Tilde{\mathbf{y}} - \Tilde{\mathbf{A}}\Tilde{\mathbf{x}}\|_2^2 
    +\lambda \sum_{i=1}^N \sqrt{(x^R_i)^2 + (x^I_i)^2},
\end{align}
where $\Tilde{\mathbf{y}} =[(\mathbf{y}^R)^\top, (\mathbf{y}^I)^\top]^\top$, $\mathbf{y}^R$ and $\mathbf{y}^I$ are the real and imaginary parts of $\mathbf{y}$ respectively, $\Tilde{\mathbf{x}} = [(\mathbf{x}^R)^\top, (\mathbf{x}^I)^\top]^\top$,
\begin{align}
    \Tilde{\mathbf{A}} = 
    \begin{bmatrix}
    \phantom{-}\mathbf{A}^R & \mathbf{A}^I\\
    -\mathbf{A}^I & \mathbf{A}^R
    \end{bmatrix},
\end{align}
and $\mathbf{A}^R$ and $\mathbf{A}^I$ are the real and imaginary parts of $\mathbf{A}$ respectively. 

Since DOA is very important in military applications, in this numerical example, we demonstrate the vulnerability of DOA estimation using group LASSO. In this experiment, we assume that there are $N=30$ sensors, $K=4$ sources, and the sources are located in the possible $M=50$ locations. The location of the $4$ sources are randomly chosen; for the real part and imaginary part of each signal, they are i.i.d. drawn from a standard normal distribution. The noise is i.i.d. distributed according to the standard Gaussian distribution with zero mean and $0.1$ standard deviation. To make our attack more practical, we only attack the measurement signal, $\mathbf{y}$. Thus, the attack process can be seen as a procedure to inject some adversarial noises into our measurements. In this attack, we set the energy of $\eta_y=1.5$ with $\ell_{\infty}$ norm constraint and set $\lambda=4 $. 
We try to suppress the source on the $(47)$th grid with arrival of angle $306^\circ$ and boost the coefficient on the $(50)$th grid that originally does not have a source target. In our experiment, we set $s_i = 20$ for $i \in S$, $e_i = -1$ for $i\in E$, $\mu_i = 20$ for $i \in U$, and step-size parameter $\alpha_t = 1/t$.

Fig.\ref{fig:doa-coffs} shows the magnitude of the original regression coefficients and the regression coefficients after attack. The non-zero coefficients exactly indicate the directions of arrival of our generated target sources. The figure demonstrates that we successfully suppressed the $(47)$th coefficient and boost the $(50)$th coefficient while keeping others almost unchanged, which successfully make the receiver believe there is no target on the $(47)$th grid and there is a counterfeit target on the $(50)$th grid. Fig.~\ref{fig:doa-y} shows the real and imaginary part of the measurements before and after our attacks. This figure reveals that, when we deliberately manipulate the regression coefficients in this example, the modified measurements just look like been perturbed by the normal noises. Hence, it is hard to detect this kind of attack.

\subsection{Attack Against Sparse Group LASSO}
In this subsection, we will use the NCEP/NCAR~Reanalysis~1 dataset \cite{kalnay1996ncep} to demonstrate our attack strategy against the sparse group LASSO based feature selection. The dataset consists of the monthly mean of temperature, sea level pressure, precipitation, relative humidity, horizontal wind speed, and vertical wind
speed from 1948 to present ($871$ months) on the globe in a $2.5^\circ\times 2.5^\circ$ resolution. For demonstration purpose, we coarse the resolution to $10^\circ\times 10^\circ$ and totally we get $403$ valid ocean locations. Our goal of this task is to analyze the dependencies between the records on the ocean and the records on a certain land. Particularly, we consider the relationship between the records on the ocean and the temperature of Brazil. Moreover, we follow  \cite{chatterjee2012sparse} to remove the seasonality and the trend in the data which may dominate the signal.

\begin{figure}[t]
    \centering
    \includegraphics[width=\linewidth]{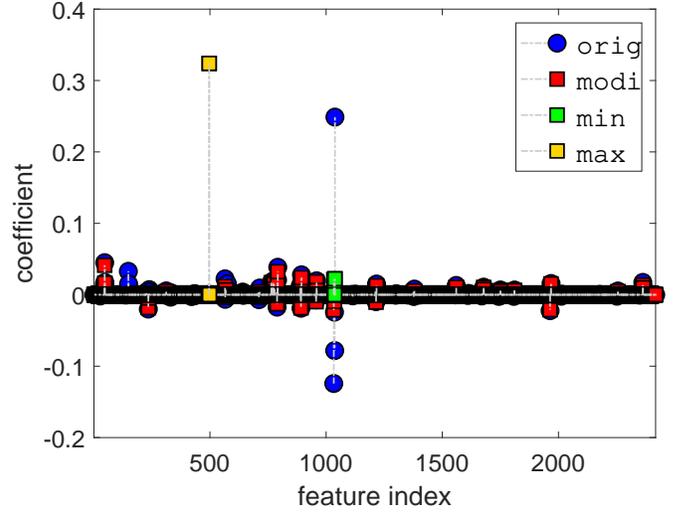}
    \caption{The regression coefficients before and after attacks. Here, `orig' denotes the original regression coefficients, `modi' represents the regression coefficients after attack,  `min' and `max' indicate the coefficients we want to supperss and boost after attack, respectively.}
    \label{fig:sp-group-lasso-coeffs}
\end{figure}

We use the data from Jan. 1984 to Dec. 2007 as the training data and the data from Jan. 2008 to Dec. 2017 as test data. Hence, we have $720$ training samples and $120$ test samples. 
We use the sparse group LASSO algorithm to find the coefficients and then use these coefficients to predict the temperature of Brazil. The regression coefficients are grouped by its location. So, each group has six coefficients. We use root mean square error (RMSE) and r-square ($r^2)$ value to measure the goodness of the regression coefficients. 
In this experiment, we set $\lambda_1=\lambda_2 = N/20$, $s_i = 1$ for $i\in S$, $e_i= -1$ for $i\in E$, $\mu_i = 20$ for $i\in U$ and $\alpha_t=1/t$. Our attack strategy is to use energy budgets $\eta_y = 0.2$ and $\eta_x=0.2$ with the $\ell_\infty$ constraints to suppress the coefficients in group $173$ and boost the coefficients in group $83$ while keeping others unchanged. 

Fig.~\ref{fig:sp-group-lasso-coeffs} depicts the coefficients before and after our attack. From the figure we can see, without attack, we can find the most representative coefficients in group $173$ with coordinate $40$W, $20$S, which is located on the ocean near the land of Brazil. 
After our attack, as demonstrated, we successfully suppressed the coefficients in group $173$ and boost the coefficients in group $83$. By doing so, it gives us the incorrect explanation of the temperature in Brazil. Further, we get $r^2 = 0.55$ and $\text{RMSE}= 0.53$ without attack on the test data. After attack, we get $r^2=0.37$ and $\text{RMSE}= 0.62$ on the test data. In summary, by attacking the training data, we can manipulate the interpretation of the relationship between the features and the response value and also worsen the prediction results.

\section{conclusion}\label{sec:conclude}
In this paper, we have investigated the adversarial robustness of the LASSO based feature selection algorithms, including ordinary LASSO, group LASSO and sparse group LASSO. We have provided an approach to mitigate the non-differentiability of the $\ell_1$ norm based feature selection methods, and have designed an algorithm to obtain the optimal attack strategy. The numerical examples both on the synthetic data and real data have shown that feature selection based on LASSO and its variants are very vulnerable to the adversarial attacks. It is of interest to study the defense strategy against this kind of attacks in the future. 

\bibliographystyle{./styles/IEEEtran}
\bibliography{mybib}

\begin{thebibliography}{10}
\providecommand{\url}[1]{#1}
\csname url@samestyle\endcsname
\providecommand{\newblock}{\relax}
\providecommand{\bibinfo}[2]{#2}
\providecommand{\BIBentrySTDinterwordspacing}{\spaceskip=0pt\relax}
\providecommand{\BIBentryALTinterwordstretchfactor}{4}
\providecommand{\BIBentryALTinterwordspacing}{\spaceskip=\fontdimen2\font plus
\BIBentryALTinterwordstretchfactor\fontdimen3\font minus
  \fontdimen4\font\relax}
\providecommand{\BIBforeignlanguage}[2]{{%
\expandafter\ifx\csname l@#1\endcsname\relax
\typeout{** WARNING: IEEEtran.bst: No hyphenation pattern has been}%
\typeout{** loaded for the language `#1'. Using the pattern for}%
\typeout{** the default language instead.}%
\else
\language=\csname l@#1\endcsname
\fi
#2}}
\providecommand{\BIBdecl}{\relax}
\BIBdecl

\bibitem{li2020advfeature}
F.~Li, L.~Lai, and S.~Cui, ``On the adversarial robustness of feature selection
  using {LASSO},'' in \emph{Proc. IEEE International Workshop on Machine
  Learning for Signal Processing}, Espoo, Finland, Sep. 2020.

\bibitem{dash1997feature}
M.~Dash and H.~Liu, ``Feature selection for classification,'' \emph{Intelligent
  Data Analysis}, vol.~1, no.~3, pp. 131--156, Jan. 1997.

\bibitem{Mandanas2020feature}
F.~D. {Mandanas} and C.~L. {Kotropoulos}, ``Subspace learning and feature
  selection via orthogonal mapping,'' \emph{IEEE Transactions on Signal
  Processing}, vol.~68, pp. 1034--1047, Jan. 2020.

\bibitem{Furlane2006feature}
C.~{Furlanello}, S.~{Merler}, and G.~{Jurman}, ``Combining feature selection
  and {DTW} for time-varying functional genomics,'' \emph{IEEE Transactions on
  Signal Processing}, vol.~54, no.~6, pp. 2436--2443, Jun. 2006.

\bibitem{tibshirani1996regression}
R.~Tibshirani, ``Regression shrinkage and selection via the {LASSO},''
  \emph{Journal of the Royal Statistical Society: Series B (Methodological)},
  vol.~58, no.~1, pp. 267--288, 1996.

\bibitem{tan2015lasso}
M.~{Tan}, I.~W. {Tsang}, and L.~{Wang}, ``Matching pursuit {LASSO} part {I}:
  Sparse recovery over big dictionary,'' \emph{IEEE Transactions on Signal
  Processing}, vol.~63, no.~3, pp. 727--741, Feb. 2015.

\bibitem{butcher2015probe}
L.~M. Butcher and S.~Beck, ``Probe lasso: a novel method to rope in
  differentially methylated regions with 450k {DNA} methylation data,''
  \emph{Methods}, vol.~72, pp. 21--28, Jan. 2015.

\bibitem{zhang2019forecasting}
Y.~Zhang, F.~Ma, and Y.~Wang, ``Forecasting crude oil prices with a large set
  of predictors: Can {LASSO} select powerful predictors?'' \emph{Journal of
  Empirical Finance}, vol.~54, pp. 97--117, Dec. 2019.

\bibitem{yang2017group}
D.~Yang and W.~Bao, ``Group lasso-based band selection for hyperspectral image
  classification,'' \emph{IEEE Geoscience and Remote Sensing Letters}, vol.~14,
  no.~12, pp. 2438--2442, Nov. 2017.

\bibitem{yuan2006model}
M.~Yuan and Y.~Lin, ``Model selection and estimation in regression with grouped
  variables,'' \emph{Journal of the Royal Statistical Society: Series B
  (Statistical Methodology)}, vol.~68, no.~1, pp. 49--67, Feb. 2006.

\bibitem{lv2011lasso}
X.~{Lv}, G.~{Bi}, and C.~{Wan}, ``The group lasso for stable recovery of
  block-sparse signal representations,'' \emph{IEEE Transactions on Signal
  Processing}, vol.~59, no.~4, pp. 1371--1382, Jan. 2011.

\bibitem{simon2013sparse}
N.~Simon, J.~Friedman, T.~Hastie, and R.~Tibshirani, ``A sparse-group
  {LASSO},'' \emph{Journal of Computational and Graphical Statistics}, vol.~22,
  no.~2, pp. 231--245, May 2013.

\bibitem{zhang2015sparsegrouplasso}
B.~{Zhang}, J.~{Geng}, and L.~{Lai}, ``Multiple change-points estimation in
  linear regression models via sparse group {LASSO},'' \emph{IEEE Transactions
  on Signal Processing}, vol.~63, no.~9, pp. 2209--2224, May 2015.

\bibitem{goodfellow2018making}
I.~Goodfellow, P.~McDaniel, and N.~Papernot, ``Making machine learning robust
  against adversarial inputs,'' \emph{Communications of the ACM}, vol.~61,
  no.~7, pp. 56--66, Jun. 2018.

\bibitem{li2020adversarial}
F.~Li, L.~Lai, and S.~Cui, ``On the adversarial robustness of subspace
  learning,'' \emph{IEEE Transactions on Signal Processing}, vol.~68, pp.
  1470--1483, Mar. 2020.

\bibitem{finlayson2019adversarial}
S.~G. Finlayson, J.~D. Bowers, J.~Ito, J.~L. Zittrain, A.~L. Beam, and I.~S.
  Kohane, ``Adversarial attacks on medical machine learning,'' \emph{Science},
  vol. 363, no. 6433, pp. 1287--1289, Mar. 2019.

\bibitem{goldblum2020adversarial}
M.~Goldblum, A.~Schwarzschild, N.~Cohen, T.~Balch, A.~B. Patel, and
  T.~Goldstein, ``Adversarial attacks on machine learning systems for
  high-frequency trading,'' \emph{arXiv:2002.09565}, Mar. 2020.

\bibitem{goodfellow6572explaining}
I.~Goodfellow, J.~Shlens, and C.~Szegedy, ``Explaining and harnessing
  adversarial examples,'' \emph{arXiv:1412.6572}, Dec. 2014.

\bibitem{balda2019adversarial}
E.~R. {Balda}, A.~{Behboodi}, and R.~{Mathar}, ``Perturbation analysis of
  learning algorithms: Generation of adversarial examples from classification
  to regression,'' \emph{IEEE Transactions on Signal Processing}, vol.~67,
  no.~23, pp. 6078--6091, Dec. 2019.

\bibitem{xiao2015feature}
H.~Xiao, B.~Biggio, G.~Brown, G.~Fumera, C.~Eckert, and F.~Roli, ``Is feature
  selection secure against training data poisoning?'' in \emph{Proc.
  International Conference on Machine Learning}, Lille, France, Jul. 2015, pp.
  1689--1698.

\bibitem{jeong2018effect}
J.~Jeong and C.~Kim, ``Effect of outliers on the variable selection by the
  regularized regression,'' \emph{Communications for Statistical Applications
  and Methods}, vol.~25, no.~2, pp. 235--243, Mar. 2018.

\bibitem{loh2011high}
P.-L. Loh and M.~J. Wainwright, ``High-dimensional regression with noisy and
  missing data: Provable guarantees with non-convexity,'' in \emph{Advances in
  Neural Information Processing Systems}, Granada, Spain, Dec. 2011, pp.
  2726--2734.

\bibitem{mei2015using}
S.~Mei and X.~Zhu, ``Using machine teaching to identify optimal training-set
  attacks on machine learners,'' in \emph{Proc. AAAI Conference on Artificial
  Intelligence}, Austin, Texas, Jan. 2015, pp. 2871--2877.

\bibitem{vincent2014sparse}
M.~Vincent and N.~R. Hansen, ``Sparse group lasso and high dimensional
  multinomial classification,'' \emph{Computational Statistics \& Data
  Analysis}, vol.~71, pp. 771--786, 2014.

\bibitem{dontchev2009implicit}
A.~L. Dontchev and R.~T. Rockafellar, ``Implicit functions and solution
  mappings,'' \emph{Springer Monographs in Mathematics. Springer}, vol. 208,
  Feb. 2009.

\bibitem{kim2007interior}
S.-J. Kim, K.~Koh, M.~Lustig, S.~Boyd, and D.~Gorinevsky, ``An interior-point
  method for large-scale $\ell_1 $-regularized least squares,'' \emph{IEEE
  Journal of Selected Topics in Signal Processing}, vol.~1, no.~4, pp.
  606--617, Dec. 2007.

\bibitem{lu2002inverses}
T.-T. Lu and S.-H. Shiou, ``Inverses of 2$\times$ 2 block matrices,''
  \emph{Computers \& Mathematics with Applications}, vol.~43, no. 1-2, pp.
  119--129, 2002.

\bibitem{condat2016fast}
L.~Condat, ``Fast projection onto the simplex and the $\ell_1$ ball,''
  \emph{Mathematical Programming}, vol. 158, no. 1-2, pp. 575--585, Sep. 2015.

\bibitem{chen2014group}
P.-Y. Chen and I.~W. Selesnick, ``Group-sparse signal denoising: non-convex
  regularization, convex optimization,'' \emph{IEEE Transactions on Signal
  Processing}, vol.~62, no.~13, pp. 3464--3478, Jul. 2014.

\bibitem{zhao2018functional}
Q.~Zhao, W.~X. Li, X.~Jiang, J.~Lv, J.~Lu, and T.~Liu, ``Functional brain
  networks reconstruction using group sparsity-regularized learning,''
  \emph{Brain Imaging and Behavior}, vol.~12, no.~3, pp. 758--770, Jun. 2018.

\bibitem{ziniel2013dynamic}
J.~Ziniel and P.~Schniter, ``Dynamic compressive sensing of time-varying
  signals via approximate message passing,'' \emph{IEEE Transactions on Signal
  Processing}, vol.~61, no.~21, pp. 5270--5284, Nov. 2013.

\bibitem{roth2008group}
V.~Roth and B.~Fischer, ``The group-lasso for generalized linear models:
  uniqueness of solutions and efficient algorithms,'' in \emph{Proc.
  International Conference on Machine Learning}, Helsinki, Finland, Jul. 2008,
  pp. 848--855.

\bibitem{chatterjee2012sparse}
S.~Chatterjee, K.~Steinhaeuser, A.~Banerjee, S.~Chatterjee, and A.~Ganguly,
  ``Sparse group {LASSO}: Consistency and climate applications,'' in
  \emph{Proc. SIAM International Conference on Data Mining}, Anaheim, CA, Apr.
  2012, pp. 47--58.

\bibitem{zhao2015heterogeneous}
L.~Zhao, Q.~Hu, and W.~Wang, ``Heterogeneous feature selection with multi-modal
  deep neural networks and sparse group {LASSO},'' \emph{IEEE Transactions on
  Multimedia}, vol.~17, no.~11, pp. 1936--1948, Nov. 2015.

\bibitem{kalivas1997two}
J.~H. Kalivas, ``Two data sets of near infrared spectra,'' \emph{Chemometrics
  and Intelligent Laboratory Systems}, vol.~37, no.~2, pp. 255--259, Jun. 1997.

\bibitem{stoica1990maximum}
P.~Stoica and K.~C. Sharman, ``Maximum likelihood methods for
  direction-of-arrival estimation,'' \emph{IEEE Transactions on Acoustics,
  Speech, and Signal Processing}, vol.~38, no.~7, pp. 1132--1143, Jul. 1990.

\bibitem{shan1985spatial}
T.-J. Shan, M.~Wax, and T.~Kailath, ``On spatial smoothing for
  direction-of-arrival estimation of coherent signals,'' \emph{IEEE
  Transactions on Acoustics, Speech, and Signal Processing}, vol.~33, no.~4,
  pp. 806--811, Aug. 1985.

\bibitem{kalnay1996ncep}
E.~Kalnay, M.~Kanamitsu, R.~Kistler, W.~Collins, D.~Deaven, L.~Gandin,
  M.~Iredell, S.~Saha, G.~White, J.~Woollen \emph{et~al.}, ``The {NCEP/NCAR}
  40-year reanalysis project,'' \emph{Bulletin of the American Meteorological
  Society}, vol.~77, no.~3, pp. 437--472, 1996.

\end{thebibliography}
\end{document}